\documentclass{kais}
\usepackage{wrapfig}
\usepackage{graphics,graphicx,amsmath,epsfig,bm,subfigure,stmaryrd}
\usepackage{amssymb}
\usepackage{url}
\usepackage{balance}
\usepackage{multirow}
\usepackage{booktabs}
 \usepackage[usenames,dvipsnames]{pstricks}
 \usepackage{epsfig}
 \usepackage{pst-grad} 
 \usepackage{pst-plot} 

\newcommand{\hcc}[0]{\textsc{Hcc}}
\newcommand{\svm}[0]{\textsc{Svm}}
\newcommand{\ica}[0]{\textsc{Ica}}
\newcommand{\cp}[0]{\textsc{Cp}}
\newcommand{\cf}[0]{\textsc{Cf}}

\newcommand{\ie}[0]{\textit{i.e.}}
\newcommand{\eg}[0]{\textit{e.g.}}
\newcommand{\etc}[0]{\textit{etc.}}

\begin{document}

\title{Meta Path-Based Collective Classification in  Heterogeneous Information Networks}

\author[X. Kong, B. Cao, P. S. Yu, Y.~Ding and David~J.~Wild]{Xiangnan Kong$^1$, Bokai Cao$^1$, Philip S. Yu$^1$, Ying~Ding$^2$ and David~J.~Wild$^2$\\
$^1$Department of Computer Science, University of Illinois at Chicago, Chicago IL, USA\\
$^2$Indiana~University~Bloomington, Bloomington, IN, USA}

\maketitle

\begin{abstract}
\textit{Collective classification} has been intensively studied due to its impact in many important applications, such as web mining, bioinformatics and citation analysis. Collective classification approaches exploit the dependencies of a group of linked objects whose class labels are correlated and need to be predicted simultaneously. In this paper, we focus on studying the collective classification problem in heterogeneous networks, which involves multiple types of data objects interconnected by multiple types of links. Intuitively, two objects are correlated if they are linked by many paths in the network. However, most existing approaches measure the dependencies among objects through directly links or indirect links without considering the different semantic meanings behind different paths. In this paper, we study the collective classification problem taht is defined among the same type of objects in heterogenous networks. Moreover, by considering different linkage paths in the network, one can capture the subtlety of different types of dependencies among objects. We introduce the concept of meta-path based dependencies among objects, where a meta path is a path consisting a certain sequence of linke types. We show that the quality of collective classification results strongly depends upon the meta paths used.  To accommodate the large network size, a novel solution, called {\hcc} (meta-path based Heterogenous Collective Classification), is developed to effectively assign labels to a group of instances that are interconnected through different meta-paths. The proposed {\hcc} model can capture different types of dependencies among objects with respect to different meta paths. Empirical studies on real-world networks demonstrate that effectiveness of the proposed meta path-based collective classification approach. 
\end{abstract}

\begin{keywords}
Heterogeneous information networks, Meta path, Collective classification
\end{keywords}


\section{Introduction}\label{sec:intro}

Collective classification methods that exploit the linkage information in networks to improve
classification accuracies have been studied intensively in the last decade.  Different from
conventional supervised classification approaches that assume data are independent and identically distributed, collective classification methods aim at exploiting the label autocorrelation among a group of inter-connected instances and predict their class labels collectively, instead of independently. In many network data \cite{TAK02,GLFW10}, the instances are inter-related with complex dependencies. For example, in bibliographic networks, if two papers both cite (or are cited by) some
other papers ({\ie}, bibliographic coupling or co-citation relationship) or one paper cites the other ({\ie}, citation
relationship), they are more likely to share similar research topics than those papers without such
relations.  These dependencies among the related instances should be considered explicitly during
classification process. Motivated by these challenges,  collective classification problem has
received considerable attention in the literature \cite{NJ00,TAK02,LG03}.

\begin{figure}
\centering
    \begin{minipage}[l]{\columnwidth}
          \centering
\includegraphics[width=1\textwidth]{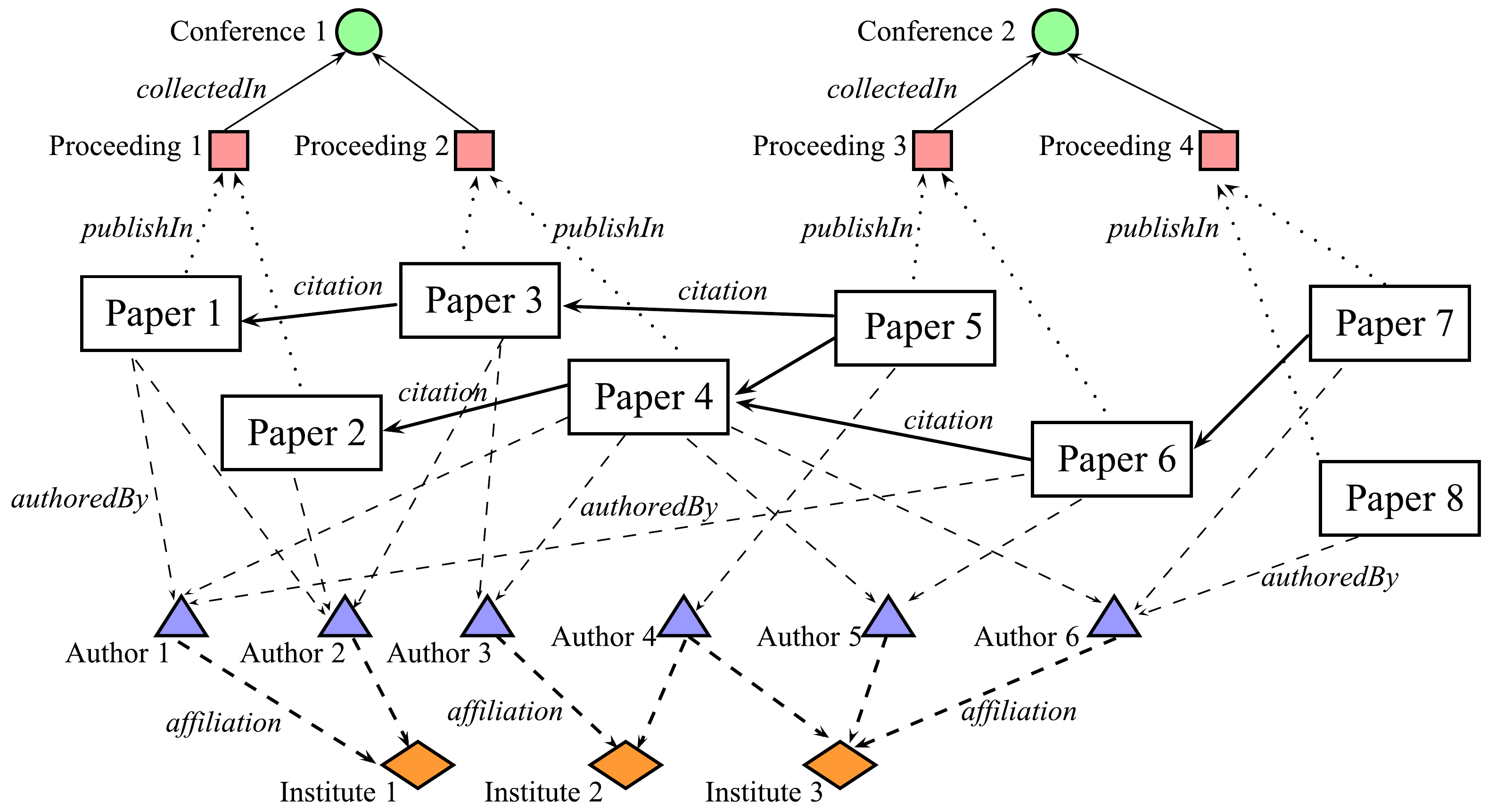}
    \end{minipage}
\caption{A Heterogeneous Information Network} \label{fig:hin_eg}
\end{figure}

Most approaches in collective classification focus on exploiting the dependencies among different interconnected objects, e.g., social networks with friendship links, webpage networks with hyper-links.  With the recent advance in data collection techniques,  many real-world applications are facing large scale heterogeneous information networks \cite{SHYYW11} with multiple types of objects inter-connected through multiple types links. These networks are multi-mode and multi-relational networks, which involves large amount of information. For example, a bibliographic network in Figure~\ref{fig:hin_eg} involves five types of nodes (papers, author, affiliations, conference and proceedings) and five types of
links. This heterogeneous information network is more complex and contain more linkage information than its homogenous sub-network, {\ie}, a paper network with only citation links.

In this paper, we focus on studying the problem of collective classification on one type of nodes
within a heterogenous information network, {\eg}, classifying the paper nodes collectively in
Figure~\ref{fig:hin_eg}. Formally, the collective classification problem in heterogeneous
information networks corresponds to predicting the labels of a group of related instances
simultaneously. Collective classification is particularly challenging in heterogenous information
networks. The reason is that, in the homogenous networks, conventional collective classification
methods can classify a group of related instances simultaneously by considering the dependencies
among instances inter-connected through one type of links. But in heterogeneous network, each
instance can have multiple types of links, and the dependencies among related instances are more
complex.

\begin{table}
\centering
\caption{Semantics of Meta Paths among Paper Nodes}\vspace{10pt}\label{tab:path_semantic}
{\scriptsize
\begin{tabular}{llll}
\toprule
&\textbf{Notation} & \textbf{Meta Path} &   \textbf{Semantics of the Dependency}\\
\midrule
1&P$\rightarrow$P & Paper$\xrightarrow{cite}$ Paper &  Citation\\
2&P$\leftarrow$P$\rightarrow$P& Paper$\xrightarrow{cite^{-1}}$ Paper $\xrightarrow{cite}$ Paper 	&  Co-citation\\
3&P$\rightarrow$P$\leftarrow$P& Paper$\xrightarrow{cite}$ Paper $\xrightarrow{cite^{-1}}$ Paper 	&  Bibliographic coupling\\
4&PVP & Paper$\xrightarrow{publishIn}$ Proceeding $\xrightarrow{publishIn^{-1}}$ Paper &  Papers in the same proceeding\\
5&PVCVP& Paper $\xrightarrow{publishIn}$ Proceeding $\xrightarrow{collectIn}$ Conference \\
&& \multicolumn{1}{r}{$\xrightarrow{collectIn^{-1}}$ Proceeding $\xrightarrow{publishIn^{-1}}$ Paper}&  Papers in the same conference\\
6&PAP& Paper$\xrightarrow{write^{-1}}$ Author $\xrightarrow{write}$ Paper &  Papers sharing authors\\
7&PAFAP & Paper$\xrightarrow{write^{-1}}$ Author$\xrightarrow{affiliation}$ Institute \\
 &&\multicolumn{1}{r}{$\xrightarrow{affiliation^{-1}}$ Author $\xrightarrow{write}$ Paper}&  Papers from the same institute\\
\bottomrule
\end{tabular}
}
\end{table}

If we consider collective classification and heterogeneous information networks as a whole, the
major research challenges can be summarized as follows:

\textbf{Multi-Mode and Multi-Relational Data:} One fundamental problem in classifying heterogeneous
information networks is the complex network structure that involves multiple types of nodes and
multiple types of links. For example, in Figure~\ref{fig:hin_eg}, one paper node can be linked
\textit{directly} with different types of objects, such as authors, conference proceedings and
other papers, through different types of links, such as \textit{citation}, \textit{authoredBy},
{\etc} Different types of links have totally different semantic meanings. Trivial application of
conventional methods by ignoring the link types and node types can not fully exploit the structural
information within a heterogeneous information network.

\textbf{Heterogeneous Dependencies:} Another problem is that objects in heterogeneous information
networks can be linked \emph{indirectly} through different types of relational paths. Each types of
relational path corresponds to different types of \emph{indirect} relationships between objects.
For example, in Figure~\ref{fig:hin_eg}, paper nodes can be linked with each other indirectly
through multiple \textit{indirect} relationships, such as, 1)  the ``paper-author-paper'' relation
indicates relationships of two papers sharing same authors; 2) the
``paper-author-institute-author-paper" relation denotes relationship between papers that are
published from the same institute. Heterogenous information  networks can encode various complex
relationships among different objects. Thus, ignoring or treating all relations equally will loss
information dependence information in a heterogeneous information network. Exploring such
heterogeneous structure information has been shown useful in many other data mining tasks, such as
ranking \cite{LC10a,LC10b}, clustering \cite{SHZYCW09,SYH09} and classification tasks \cite{JHD11}.

In this paper, we study the problem of collective classification in heterogeneous information
networks and propose a novel solution, called {\hcc} (meta-path based Heterogenous Collective Classification), to
effectively assign class labels to one type of objects in the network. Different from conventional
collective classification methods, the proposed {\hcc} model can exploit a large number of
different types of dependencies among objects simultaneously. We define meta path-based
dependencies to capture different types of relationships among objects. By explicitly exploiting
these dependencies, our {\hcc} method can effectively exploit the complex relationships among
objects. Empirical studies on real-world tasks demonstrate that the proposed approach can
significantly boost the collective classification performances in heterogeneous information networks.

The rest of the paper is organized as follows. We start by a brief review on related work of
collective classification and heterogeneous information networks. Then we introduce the preliminary
concepts, give the problem definitions in Section~\ref{sec:prob_def} and present the {\hcc}
algorithm in Section~\ref{sec:hcc}. Then Section~\ref{sec:experiment} reports the experiment
results. In Section~\ref{sec:conclusion}, we conclude the paper.

\section{Related Work}\label{sec:relatedwork} 

Our work is related to both collective classification techniques on relational data and heterogeneous information
networks. We briefly discuss both of them.

Collective classification of relational data has been investigated by many researchers. The task is
to predict the classes for a group of related instances simultaneously, rather than predicting a
class label for each instance independently. In relational datasets, the class label of one instance
can be related to the class labels (sometimes attributes) of the other related instances. Conventional
collective classification approaches focus on exploiting the correlations among the class labels of
related instances to improve the classification performances. Roughly speaking, existing collective
classification approaches can be categorized into two types based upon the different approximate
inference strategies: (1) Local methods: The first type of approaches employ a local classifier to
iteratively classify each unlabeled instance using both attributes of the instances and relational
features derived from the related instances. This type of approaches involves an iterative process
to update the labels and the relational features of the related instances, {\eg} iterative
convergence based approaches \cite{NJ00,LG03} and Gibbs sampling approaches \cite{MGA09}.
Many local classifiers have been used for local methods, {\eg} logistic regression \cite{LG03},
Naive Bayes \cite{NJ00}, relational dependency network \cite{NJ03}, {\etc} (2) Global methods: The
second type of approaches optimizes global objective functions on the entire relational dataset,
which also uses both attributes and relational features for inference \cite{TAK02}.
For a detailed review of collective classification please refer to \cite{SNBG08}.

Heterogeneous information networks are special kinds of information networks which involve multiple
types of nodes or multiple types of links.
In a heterogeneous information network, different types of nodes and edges have different semantic
meanings. The complex and semantically enriched network possesses great potential for knowledge
discovery. In the data mining domain, heterogeneous information networks are ubiquitous in many
applications, and have attracted much attention in the last few years \cite{SYH09,SHZYCW09,JHD11}. 
Sun et al. \cite{SYH09,SHYYW11} studied the clustering problem and top-k similarity problem in
heterogeneous information networks. Ming et al. studied a specialized classification problem on
heterogeneous networks, where different types of nodes share a same set of label concepts
\cite{JHD11}. However, these approaches are not directly applicable in collective classification
problems, since focus on convention classification tasks without exploiting the meta path-based
dependencies among objects. 

\section{Problem Definition}\label{sec:prob_def}
In this section, we first introduce several related concepts and notations. Then, we will formally
define the collective classification problem in heterogeneous information networks.. 

\smallskip
\textsc{Definition 1}. \textbf{Heterogeneous Information Network}:
A heterogeneous information network \cite{SYH09,SHYYW11} is a special kind of information network,
which is represented as a directed graph $\mathcal{G}=(\mathcal{V},\mathcal{E})$. $\mathcal{V}$ is
the set of nodes, including $t$ types of objects $\mathcal{T}_1= \left\{v_{11}, \cdots, v_{1
n_1}\right\}, \cdots , \mathcal{T}_t= \{v_{t1}, \cdots, v_{t n_t}\}$. $\mathcal{E}\subseteq
\mathcal{V} \times \mathcal{V}$ is the set of links between the nodes in $\mathcal{V}$, which
involves multiple types of links.

\smallskip
\textsc{Example 1}. \textbf{ACM conference network}:
A heterogeneous information network graph is provided in Figure~\ref{fig:hin_eg}. This network
involves five types of objects, {\ie}, papers (P), authors (A), institutes (F), proceedings (V) and
conferences (C), and five types of links, {\ie}, \emph{citation}, \emph{authoredBy},
\emph{affiliation}, \emph{publishedIn} and \emph{collectedIn}.
\smallskip

\begin{table}
    \centering
    \caption{Important Notations.}\label{tab:notation}
    \begin{tabular}{|r|l|} 
    \toprule
	Symbol& Definition\\ 
    \midrule
    $\mathcal{V}=\bigcup_{i=1}^t \mathcal{T}_i$ & the set of nodes, involving $t$ types of nodes\\ 
    $\mathcal{E}=\{e_i\in \mathcal{V}\times\mathcal{V}\}$  &the set of edges or links\\ 
    $\mathcal{X}=\{\bm{x}_1, \cdots, \bm{x}_{n_1}\}$ & the given attribute values for each node in target type $\mathcal{T}_1$\\ 
    $\mathcal{Y}= \left\{Y_1, \cdots, Y_{n_1}\right\}$   &the set of variables for labels of the nodes in $\mathcal{T}_1$, and $Y_i\in\mathcal{C}$\\ 
    $\mathcal{L}$ and $\mathcal{U}$ & the sets for training nodes and testing nodes, and $\mathcal{L}\cup \mathcal{U} = \mathcal{T}_1$\\ 
    $ y_i $      &  the given label for node $v_{1i} \in \mathcal{L}$, and $Y_i =y_i$\\ 
    $\mathcal{S}=\{\mathcal{P}_1, \cdots, \mathcal{P}_m\}$ & the set of meta paths\\ 
    $\mathcal{P}_{j}(i) =\left\{ k | \mathcal{P}_j (v_{1i}, v_{1k})\right\}$  &  the index set of all related instances to $\bm{x}_i$ through meta path $\mathcal{P}_j$\\ 
     \bottomrule
    \end{tabular} 
\end{table}

Different from conventional networks, heterogeneous information networks involve different types of
objects (e.g., papers and conference) that are connected with each other through multiple types of
links. 
Each type of links represents an unique binary relation $R$ from node type $i$  to node type $j$,
where $R(v_{ip},v_{jq})$ holds iff object $v_{ip}$ and $v_{jq}$ are related by relation $R$.
$R^{-1}$ denotes the inverted relation of $R$, which holds naturally for $R^{-1}(v_{jq},v_{ip})$.
Let $dom(R) = \mathcal{T}_i$ denote the domain of relation $R$, $rang(R)= \mathcal{T}_j$ denotes
its range.  $R(a)=\{b : R(a,b)\}$.  For example, in Figure~\ref{fig:hin_eg}, the link type
``\emph{authorBy}" can be written as a relation $R$ between paper nodes and author nodes.
$R(v_{ip},v_{jq} )$ holds iff author $v_{jq}$ is one of the authors for paper $v_{ip}$.  For
convenience, we can write this link type as ``$paper \xrightarrow{authoredBy} author$" or
``$\mathcal{T}_i \xrightarrow{R}\mathcal{T}_j$".

In heterogenous information networks, objects are also inter-connected through indirect links,
{\ie}, paths. For example, in Figure~\ref{fig:hin_eg}, paper 1 and paper 4 are linked through a
sequence of edges: ``$paper 1\xrightarrow{authoredBy} author 1 \xrightarrow{authoredBy^{-1}} paper
4$". In order to categorize these paths, we extend the definition of link types to ``path types",
which are named as \emph{meta path}, similar to \cite{SHYYW11,LC10a}.

\smallskip
\textsc{Definition 2}. \textbf{Meta Path}:
A meta path $\mathcal{P}$ represents a sequence of relations $R_1,\cdots,R_\ell$ with constrains
that $\forall i\in\{1,\cdots,\ell-1\}, rang(R_i )=dom(R_{i+1} )$. The meta path $\mathcal{P}$ can
also be written as
$\mathcal{P}:\mathcal{T}_1\xrightarrow{R_1}\mathcal{T}_2\xrightarrow{R_2}\cdots\xrightarrow{R_\ell}\mathcal{T}_{\ell+1}$,
{\ie}, $\mathcal{P}$ corresponds to a composite relation $R_1 \times R_2 \times \cdots \times
R_\ell$ between node type $\mathcal{T}_1$ and $\mathcal{T}_{\ell+1}$. $dom(P)=dom(R_1)$ and
$rang(P)=rang(R_\ell )$. The length of $P$  is $\ell$, {\ie}, the number of relations in
$\mathcal{P}$.
\smallskip

Different meta paths usually represent different semantic relationships among linked objects. In
Table~\ref{tab:path_semantic}, we show some examples of meta paths with their corresponding
semantics.  Most conventional  relationships studied in network data can naturally be captured by
different meta paths. For example, the paper \emph{co-citation} relation \cite{DYFC09} can
naturally be represented by meta path
``$paper\xrightarrow{cite^{-1}}paper\xrightarrow{cite}paper$", and the co-citation frequencies can
be written as the number of path instances for the meta path. Here a \emph{path instance} of
$\mathcal{P}$, denoted as $p\in \mathcal{P}$, is an unique sequence of nodes and links in the
network that follows the meta path constrains. For convenience, we  use the node type sequence to
represent a meta path, {\ie}, $\mathcal{P}= \mathcal{T}_1 \mathcal{T}_2 \cdots \mathcal{T}_{l+1}$.
For example, we use $PAP$ to represent the meta path ``$paper\xrightarrow{authoredBy}
author\xrightarrow{authoredBy^{-1}} paper$". Note that for meta paths involving \emph{citation}
links, we explicitly add arrows to represent the link directions, {\eg}, the paper
\emph{co-citation} path can be written as $P\leftarrow$$P\rightarrow$$P$.

{\smallskip}
\noindent\textbf{Collective Classification in Heterogeneous Information Networks}{\smallskip}\\
\noindent
In this paper, we focus on studying the collective classification problem on \emph{one} type of
objects, instead of on all types of nodes in heterogeneous information networks. This problem
setting exists in a wide variety of applications. The reasons are as follows: in heterogenous
information networks, the label space of different types of nodes  are quite different, where we
can not assume all types of node share the same set of label concepts. For example, in medical
networks, the label concepts for patient classification tasks are only defined on patient nodes,
instead of doctor nodes or medicine nodes. In a specific classification task, we usually only care
about the classification results on one type of node. Without loss of generality, we assume the
node type $\mathcal{T}_1$ is the target objects  we need to classify. Suppose we have $n$ nodes in
$\mathcal{T}_1$.  On each node $v_{1i}\in \mathcal{T}_1$, we have a vector of attributes $\bm{x}_i
\in \mathbb{R}^d$ in the $d$-dimensional input space, and  $\mathcal{X}=\{\bm{x}_1, \cdots,
\bm{x}_{n_1}\}$. Let $\mathcal{C}=\{c_1, c_2, \cdots, c_q\}$ be the $q$ possible class labels. On
each node $v_{1i}\in \mathcal{T}_1$, we also have a label variable $Y_i \in \mathcal{C}$ indicating
the class label assigned to node $v_{1i}$, $\mathcal{Y}=\{Y_i\}_{i=1}^{n_1}$.

Assume further that we are given a set of known values $\mathcal{Y}_{L}$ for nodes in a training
set $\mathcal{L} \subset \mathcal{T}_1$, and $L$ denotes the index set for training data.
$\mathcal{Y}_{L}=\left\{y_i |  i \in L \right\}$, where $y_i \in \mathcal{C}$ is the observed
labels assigned to node $x_{1i}$.  Then the task of collective classification in heterogeneous
information networks is to infer the values of $Y_i \in \mathcal{Y}_U$ for the remaining nodes in
the testing set ($\mathcal{U}= \mathcal{T}_1 - \mathcal{L}$).

\begin{figure}
    	\begin{minipage}[l]{0.4\columnwidth}
          	\centering
		\includegraphics[width=1\textwidth]{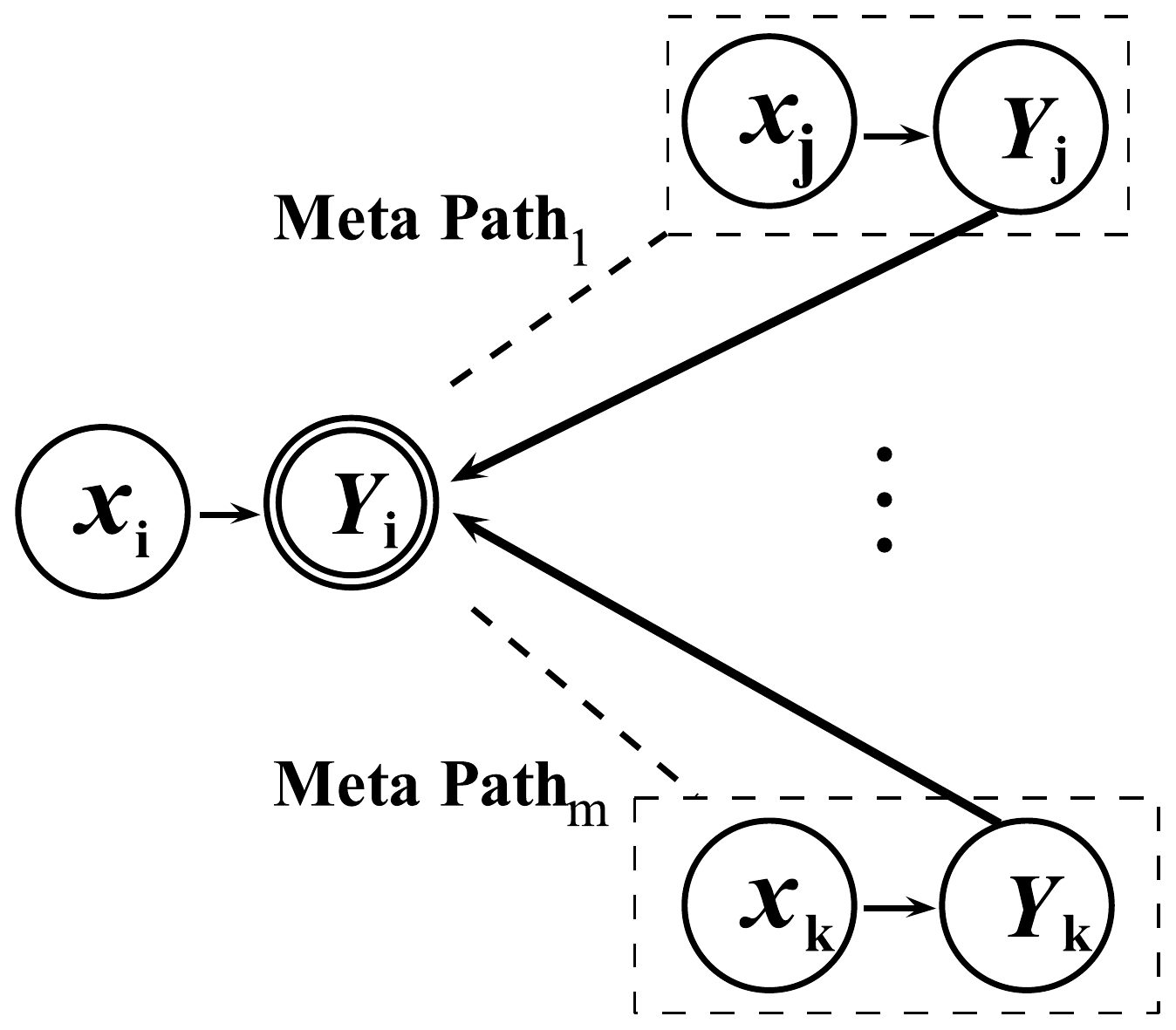}
	\end{minipage}
\caption{Meta path-based dependencies in collective classification 
for heterogeneous information networks. $Y_i$ with double circles denotes 
the current label variable to be predicted. Each rectangle represents a 
group of instances following the same meta path. 
$\bm{x}_i$ denotes the attribute values of the instance. } \label{fig:dependency}
\end{figure}

As reviewed in Section~\ref{sec:relatedwork}, the inference problem in classification tasks is to
estimate $Pr(\mathcal{Y}|\mathcal{X})$ given a labeled training set. Conventional classification
approaches usually require i.i.d. assumptions, the inference for each instance is performed independently:
$$
	Pr(\mathcal{Y}|\mathcal{X}) \propto \prod\limits_{i\in U} Pr(Y_i|\bm{x}_i)
$$
{\smallskip}
\noindent\textbf{Homogeneous Link-based Dependency}{\smallskip}\\
\noindent
In collective classification problems, the labels of related instances are not independent, but are
closely related with each other.  Conventional approaches focus on exploiting label dependencies
corresponding to one types of  homogeneous links to improve the classification performances, e.g.,
citation links in paper classification tasks, co-author links in expert classification tasks. 
These methods can model $ Pr(Y_i | \bm{x}_i, \bm{Y}_{\mathcal{P}(i)}) $. Here
$\bm{Y}_{\mathcal{P}(i)}$ denotes the vector containing all variable $Y_j$ ($\forall j\in
\mathcal{P}(i)$), and $\mathcal{P}(i)$ denotes the index set of related instances to the $i$-th
instance through meta path $\mathcal{P}$. Hence, by considering the single type of dependencies, we
will have $$ Pr(\mathcal{Y}|\mathcal{X}) \propto \prod\limits_{i\in U} Pr(Y_i| \bm{x}_i, \bm{Y}_{\mathcal{P}(i)}) $$

{\smallskip}
\noindent\textbf{Meta Path-based Dependency}{\smallskip}\\
\noindent
In heterogeneous information networks, there are complex dependencies not only among instances
directly linked through links, but also among instances indirectly linked through different meta
paths. In order to solve the collective classification problem more effectively, in this paper, we
explicitly consider different types of meta-path based dependencies in heterogeneous information
networks. Meta path-based dependences refer to the dependencies among instances that are
inter-connected through a meta path.

To the best of our knowledge, meta path-based dependencies have not been studied in collective
classification research before. Given a set of meta paths $\mathcal{S}=\{\mathcal{P}_1, \cdots,
\mathcal{P}_m\}$, the meta path-based dependency models  are
shown in Figure~\ref{fig:dependency}, {\ie},  $ Pr(Y_i | \bm{x}_i,
\bm{Y}_{\mathcal{P}_1(i)}, \bm{Y}_{ \mathcal{P}_2(i)}, \cdots, \bm{Y}_{ \mathcal{P}_m(i)}) $. $\mathcal{P}_j(i)$ denotes the index set of related instances
to the $i$-th instance through meta path $\mathcal{P}_j$.

\begin{figure}
    \begin{minipage}[l]{\columnwidth}
          \centering
\includegraphics[width=1\textwidth]{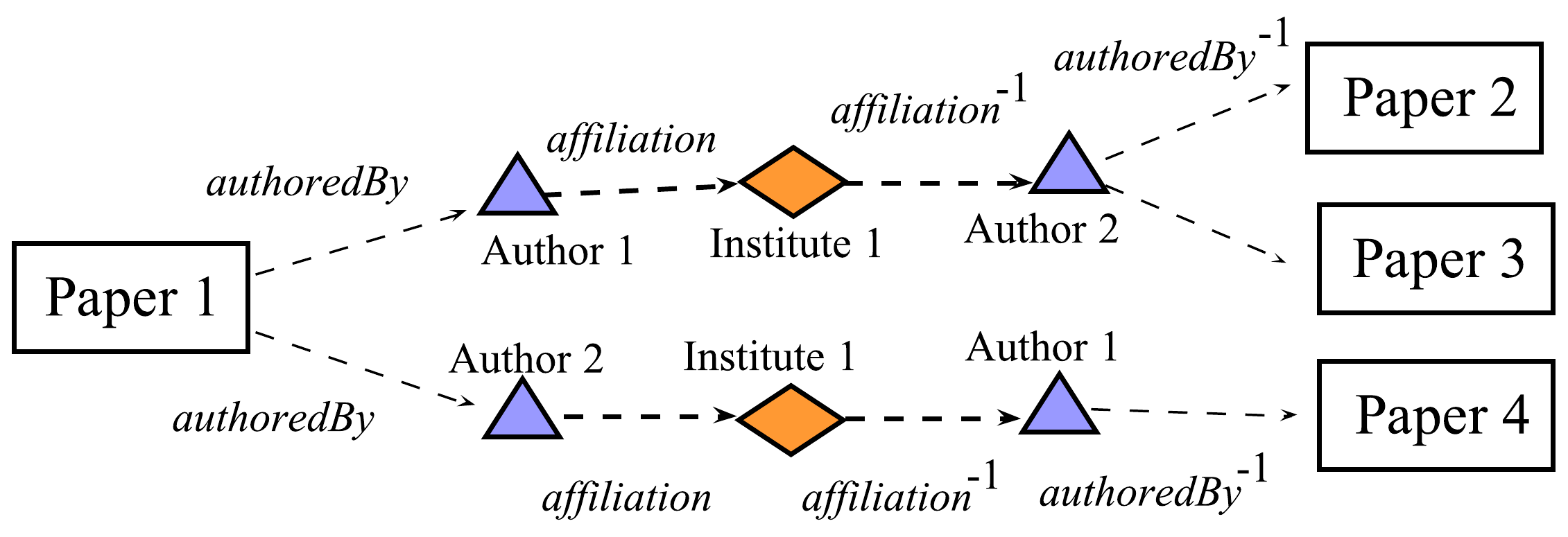}
    \end{minipage}
\caption{Path instances corresponding to the meta path $PAFAP$.} \label{fig:path_eg}
\end{figure}

For each meta path, one instance can be connected with multiple related instances in the network.
For example, in Figure~\ref{fig:path_eg}, Paper 1 is correlated with Paper 2, 3 and 4 through meta
path $\mathcal{P}_i = PAFAP$, {\ie}, $\mathcal{P}_i (Paper 1) =\{ Paper\ 2, 3, 4\}$. Hence, by
considering meta path-based dependencies, we will have $$ Pr(\mathcal{Y}|\mathcal{X}) =
\prod\limits_{i\in U} Pr\left(Y_i| \bm{x}_i, \bm{Y}_{ \mathcal{P}_1(i)}, \bm{Y}_{
\mathcal{P}_2(i)}, \cdots, \bm{Y}_{ \mathcal{P}_m(i)}\right) $$

\section{Meta Path-based Collective Classification}\label{sec:hcc}

For classifying target nodes in a heterogeneous information network, the most na\"ive approach is
to approximate $Pr(\mathcal{Y}|\mathcal{X}) \propto\prod\nolimits_{i\in U} Pr(Y_i|\bm{x}_i)$ with
the assumptions that all instances are independent from each other. However, this approach can be
detrimental to their performance for many reasons. This is particularly troublesome when nodes in
heterogeneous networks have very complex dependencies with each other  through different meta paths.

In this section, we propose a simple and effective algorithm for meta path-based collective
classification in heterogeneous information networks. We aim to develop a model to estimate the probabilities
$Pr\left(Y_i| \bm{x}_i, \bm{Y}_{\mathcal{P}_1(i)}, \cdots, \bm{Y}_{\mathcal{P}_m(i)}\right)$. We first introduce how the extract the set of meta paths from a heterogeneous
information network, then propose our collective classification algorithm, called {\hcc}
(Heterogeneous Collective Classification).

\begin{figure}[t]
    \begin{minipage}[l]{\columnwidth}
    \centering
\includegraphics[width=1\textwidth]{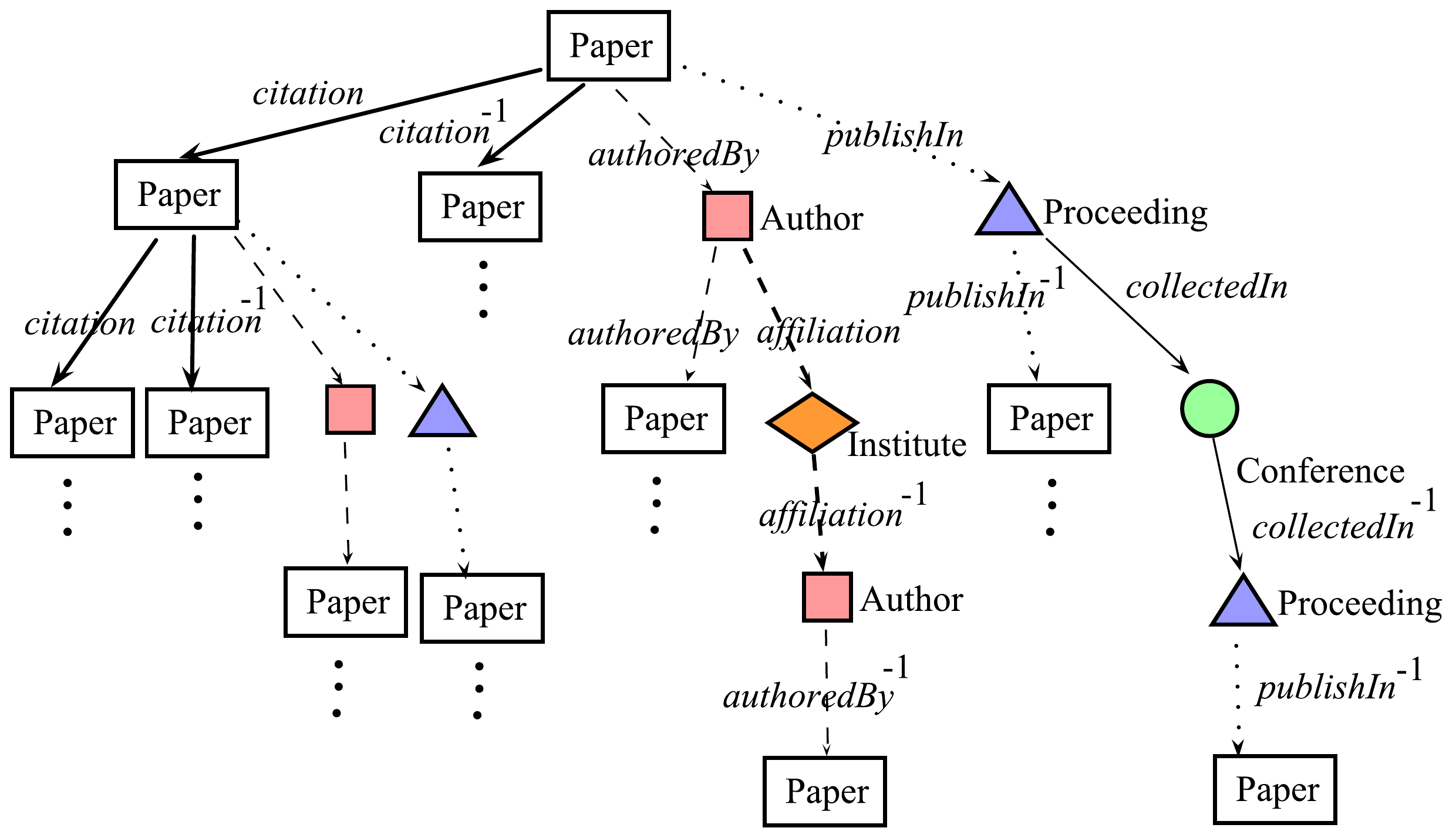}
    \end{minipage}
\caption{An example of dependence tree for meta path-based dependencies. Each paper node corresponds to a unique type of path-based dependencies in the network.} \label{fig:prefix}
\end{figure}

We first consider how to extract all meta paths in a heterogeneous information network of bounded
length $\ell_{max}$. When $\ell_{max}$ is small, we can easily generate all possible meta paths as
follows: We can organize all the type-correct relations into a prefix tree, called \emph{dependence
tree}. In Figure~\ref{fig:prefix}, we show an example of dependence tree in ACM conference
networks. The target nodes for classification are the paper nodes, and each paper node in the
dependence tree corresponds to an unique meta path, indicating one type of dependencies among paper
instances.  However, in general the number of meta paths grows exponentially with the maximum path
length $\ell_{max}$.  As it has been showed in \cite{SHYYW11}, long meta paths may not be quite
useful in capturing the linkage structure of heterogeneous information networks. In this paper, we
only exploit the instance dependences with short meta paths ($\ell_{max}=4$).

In many really world network data, exhaustively extracting all meta paths may result in large
amount of redundant meta paths, {\eg}, $PVPVP$. Including redundant meta paths in a collective
classification model can result in overfitting risks, because of additional noisy features. Many of
the redundant meta paths are constructed by combining two or more meta paths, {\eg}, meta path
$PVPVP$ can be constructed by two $PVP$ paths. In order to reduce the model's overfitting risk, we
extract all meta paths that cannot be decomposed into shorter meta paths (with at least one
\textit{non-trivial} meta paths). Here non-trivial meta paths refer to the paths with lengths
greater than 1. For example, in ACM conference network, meta paths like $P$$\rightarrow$$PAP$ can
be decomposed into $P$$\rightarrow$$P$ and $PAP$, thus will be excluded from our meta path set. In
Figure~\ref{fig:algorithm}, we the meta path set extract process as the ``Initialization" step of
our proposed method. By breadth-first search on the dependence tree, our model first select
shortest meta paths from the network. Then longer meta paths are incrementally selected into path
set $\mathcal{S}$ until we reach a meta path that can be decomposed into shorter meta paths in
$\mathcal{S}$.

After the meta path set $\mathcal{S}$ is extracted from the heterogeneous information network, we
then show how to use these meta paths to perform collective classification effectively.
Conventional collective classification based on iterative inference process, {\eg} ICA (Iterative
Classification Algorithm) \cite{NJ00,LG03}, provide a simple yet very effective method for
collective classification in homogeneous networks. Inspired by the success of these iterative
inference methods, in this paper, we propose a similar framework for meta path-based collective
classification method. This approach is called {\hcc} (Heterogeneous Collective Classification),
summarized in Figure~\ref{fig:algorithm}.

\begin{figure}
\center
\small
\begin{tabular}{l}
\toprule
  \textbf{Input:}\\
  \begin{tabular}{r@{: }l@{ }r@{: }l}
	$\mathcal{G}$ & a heterogeneous information network, &$\ell_{max}$ & maximum meta path length.\\
	$\mathcal{X}$ & attribute vectors for all target instances, &$\mathcal{Y}_L$ & labels for the training instances.\\
	$L$ & the index set for training instances, & $U$ & the index set for testing instances.\\
	A&a base learner for local model, & $Max\_It$ & maximum \# of iteration.\\
  \end{tabular}\\
\textbf{Initialize:}\\
\quad - Construct the meta path set $\mathcal{S}=\{\mathcal{P}_1,\cdots, \mathcal{P}_m\}$ by searching the dependence tree on $\mathcal{G}$:\\ 
\quad\quad Breadth first search on dependence tree by adding short meta paths into $\mathcal{S}$ first:\\ 
\quad\quad\quad 1. If the length of meta path in current tree node is greater than $\ell_{max}$, exit the BFS;\\
\quad\quad\quad 2. If the current meta path $\mathcal{P}_j$ in current tree node cannot be reconstructed by the paths in $\mathcal{S}$,\\
\quad\quad\quad\quad Add $\mathcal{P}_j$ into $\mathcal{S}$; Otherwise, prune the current node from BFS.\\

\textbf{Training:}\\
\quad - Learn the local model $f$:\\  
\quad\quad 1. Construct an extended training set $\mathcal{D}=\left\{(\bm{x}_i', \bm{y}_i)\right\}$ by converting each instance $\bm{x}_i$ to $\bm{x}_i'$ as follows:\\ 
\quad\quad\qquad $\bm{x}_i' = \left( \bm{x}_i, \text{PathRelFeature}(v_{1i}, \mathcal{E}_L, \mathcal{Y}_L, \mathcal{S})\right)$\\ 
\quad\quad 2. Let $f = A(\mathcal{D})$ be the local model trained on $\mathcal{D}$.\\
\textbf{Bootstrap:}\\
\quad - Estimate the labels, for $i\in U$\\
     \qquad 1. Produce an estimated value $\hat{Y}_i$ for $Y_i$ as follows:\\
     \qquad \qquad $\hat{Y}_i$ = $f\left((\bm{x}_i, \bm{0}) \right)$ using attributes only.\\
\textbf{Iterative Inference:}\\
\quad - Repeat until convergence or \#iteration$> Max\_It$\\
\quad \quad 1. Construct the extended testing instance by converting each instance $\bm{x}_i$ to $\bm{x}_i'$ ($ i \in U$) as follows:\\
\quad \quad \qquad  $\bm{x}_i'= \left(\bm{x}_i, \text{PathRelFeature}(v_{1i},\mathcal{E}, \mathcal{Y}_L \cup \{\hat{Y}_i|i\in U\}, \mathcal{S})\right)$\\ 
\quad \quad 2. Update the estimated value $\hat{Y}_i$ for $Y_i$ on each testing instance ($i\in U$) as follows:\\
 \qquad \qquad $\hat{Y}_i$ = $f$($\bm{x}_i'$).\\
  \textbf{Output:}\\
  \begin{tabular}{rl}
      $\hat{\bm{Y}}_U = \left(\hat{Y}_1, \cdots, \hat{Y}_n\right)$:& The labels of test instances ($i\in U$).
    \end{tabular}\\
    \bottomrule
  \end{tabular}
  \caption{The {\hcc} algorithm}\label{fig:algorithm}
\end{figure}

The general idea is as follows: we model the joint probability based
on the following assumption: if instance $v_{1i}$ and $v_{1j}$ are
not connected via any meta path in $\mathcal{S}$, the variable $Y_i$ is
conditional independent from $Y_j$ given the labels of all
$v_{1i}$'s related instances, {\ie}, $ \{v_{1j} | j\in \bigcup_{k=1}^m \mathcal{P}_k(i) \}$. Hence
the local conditional probability each instance's label can be modeled by a base learner with
extended \textit{relational features} built upon the predicted $Y_j$'s ($j\in \bigcup_{k=1}^m
\mathcal{P}_k(i)$). And the joint probability can be modeled based on these local conditional
probabilities by treating the instances as being independent.

\begin{figure}
\center
\begin{tabular}{l}
\toprule
 $\bm{x}^r=\text{PathRelFeature}\left(v,\mathcal{E}, \{Y_i\}, \mathcal{S}=\{\mathcal{P}_1, \cdots, \mathcal{P}_m\}\right)$\\
 \midrule
\qquad For each meta path $\mathcal{P}_i \in \mathcal{S}$:\\
\qquad\qquad 1. Get related instances  $\mathcal{C}=\mathcal{P}_i (v, \mathcal{E})$\\
\qquad\qquad 2. $\bm{x}^i= Aggregation\left( \{Y_j| v_{1j}\in \mathcal{P}_i(v)\} \right)$\\
\qquad Return relational feature $\bm{x}^r =\left(\bm{x}^1, \cdots, \bm{x}^m \right)$\\
\bottomrule
  \end{tabular}
  \caption{Constructing meta path-based relational features (PathRelFeature). }\label{alg:relfeature}
\end{figure}

In collective classification, each instance may be linked with different number of instances
through one meta path. In order to build a fixed number of relational features for each instance,
we employs \textit{aggregation} functions to combine the predictions on the labels of related
instances. Many aggregation functions can be used here, such as COUNT and MODE aggregators
\cite{LG03}. In this paper, we use the \emph{weighted label fraction} of the related instances as
the relational feature for each meta path. We calculate the average fraction of each label
appearing in the related instances. Each related instance in re-weighted by the number of path
instances  between from the current node, {e.g.}, for meta path $PAP$, the papers that share more
authors in their author lists are more likely to share similar topics than those only share one
author. In detail, given an aggregation function, we can get one set of relational features from
the labels of related instances for each meta path, as shown in Figure~\ref{alg:relfeature}.

Inspired by the success of ICA framework \cite{LG03,MGA07,MGA09} in collective classification, we
designed a similar inference procedure for our {\hcc} method as shown in
Figure~\ref{fig:algorithm}. (1) For inference steps, the labels of all the unlabeled instances are
unknown. We first \textit{bootstrap} an initial set of label estimation for each instance using
content attributes of each node. In our current implementation, we simply set the relational
features of unlabeled instances with zero vectors. Other strategies for \textit{bootstrap} can also
be used in this framework. (2)  \textit{Iterative Inference}: we iteratively update the relational
features based on the latest predictions and  then  these new features are used to update the
prediction of local models on each instance. The iterative process terminates when convergence
criteria are met. In our current implementation, we update the variable $Y_i$ in the $(r+1)$-th
iteration ( say $\hat{Y}_i^{(r+1)}$) using the predicted values in the $r$-th iteration
($\hat{Y}_j^{(r)}$) only.

\begin{table}
    \centering
    \caption{Summary of experimental datasets. }\label{tab:datastat}
    \begin{tabular}{lrrrr}
        \toprule
            & \multicolumn{4}{c}{Data Sets}\\
            \cmidrule{2-5}
            Characteristics
            & ACM-A
            & ACM-B
            & DBLP
            & SLAP\\
        \midrule
        \# Feature	&1,903 &376   &1,618 & 3,000 \\
        \# Instance &12,499 &10,828  &4,236 & 3714 \\
        \# Node Type    &5 &5  &2& 10\\ 
        \# Link Type     &5 &5  &2& 11\\ 
         \# Class 	&11	&11&4 & 306\\
        \bottomrule
    \end{tabular}
\end{table}


\section{Experiments}\label{sec:experiment}
{\smallskip}

\subsection{Data Collection}

In order to validate the collective classification performances, we tested our algorithm on four 
real-world heterogeneous information networks (Summarized in Table~\ref{tab:datastat}).

\begin{figure}
  \centering
  \subfigure[ACM Conference Datasets]{
    \begin{minipage}[l]{0.48\columnwidth}
      \centering
\includegraphics[width=1\textwidth]{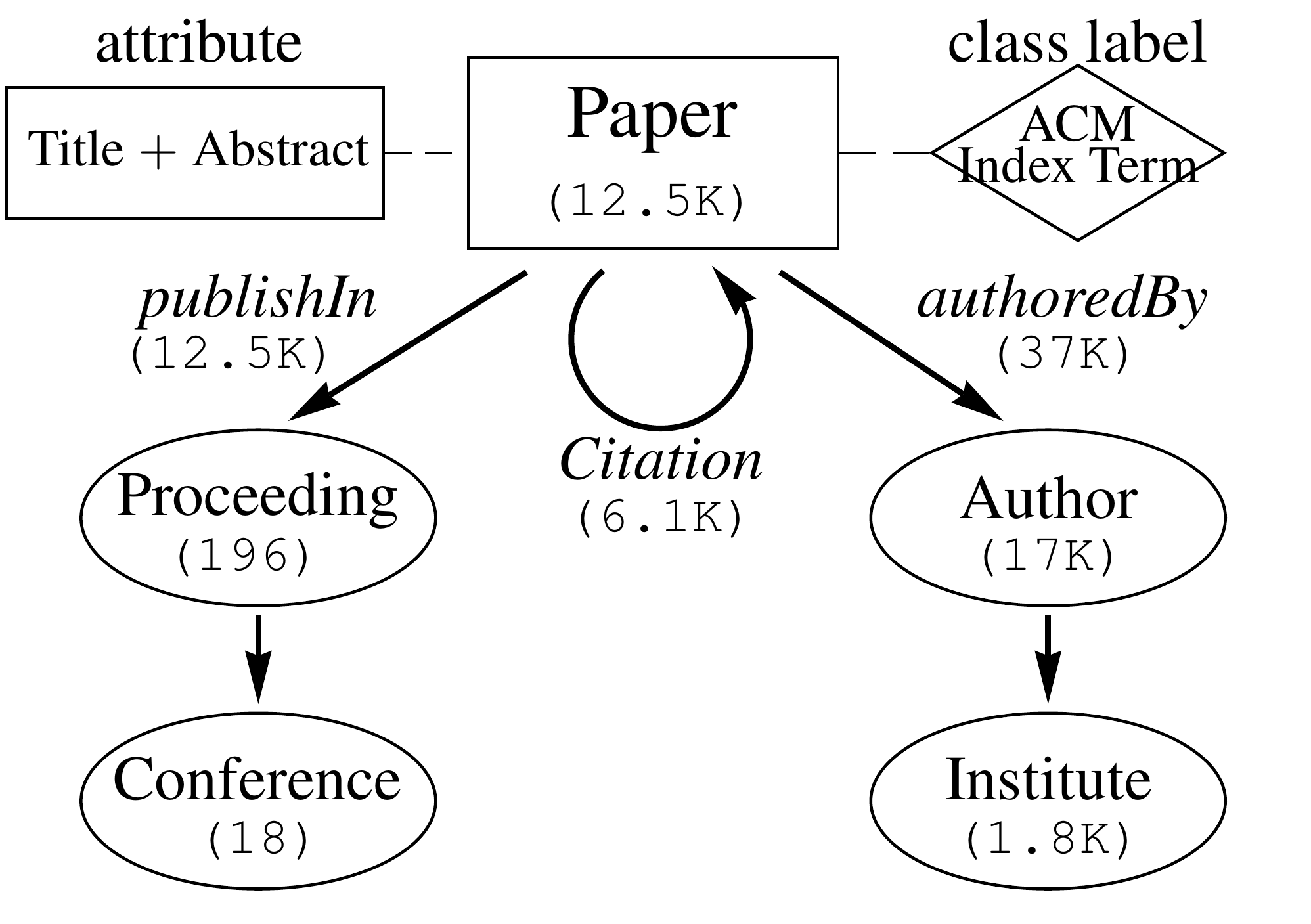}
    \end{minipage}\label{fig:acmschema}
  }
   \subfigure[DBLP Dataset]{ 
     \begin{minipage}[l]{0.48\columnwidth}
       \centering
\includegraphics[width=1\textwidth]{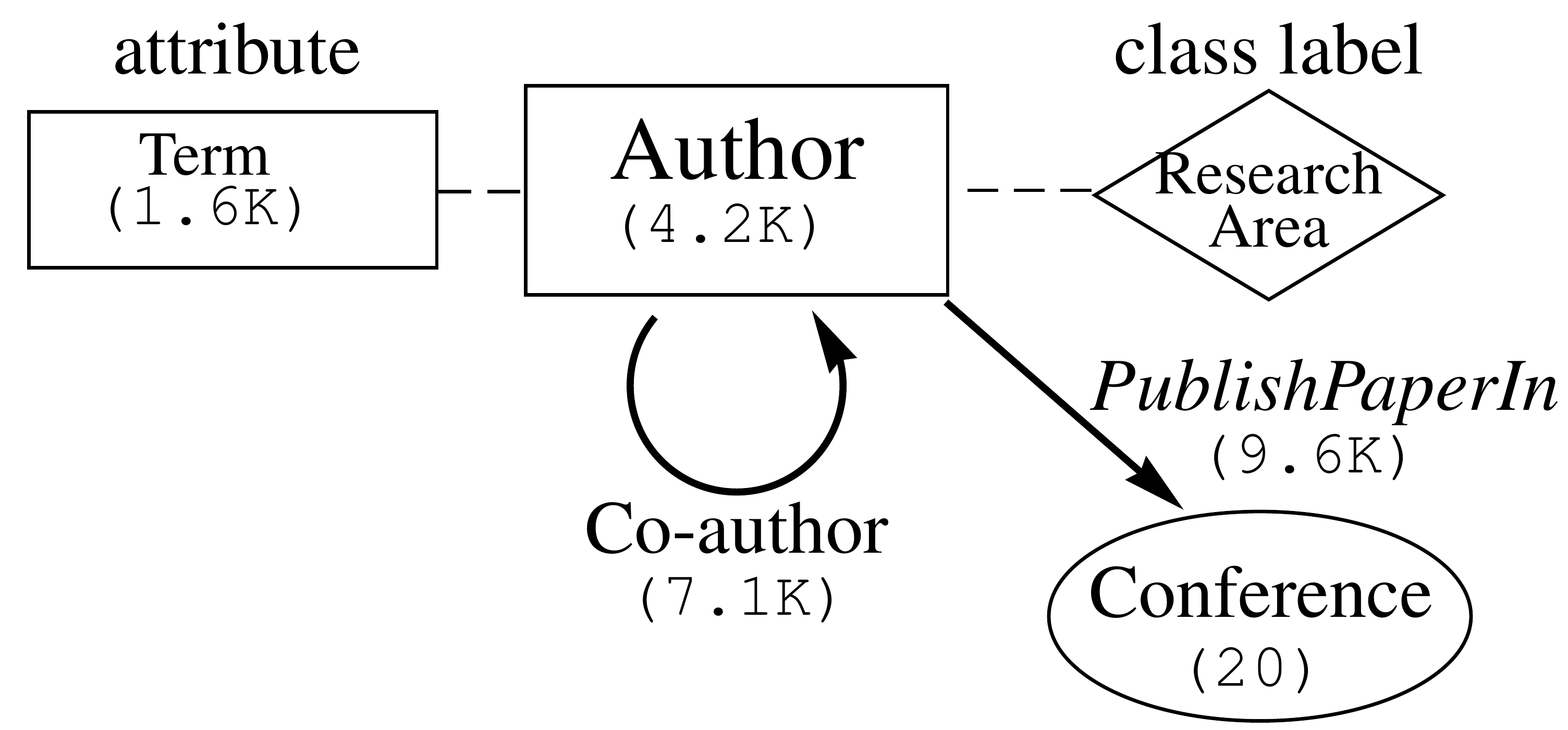}
      \end{minipage}\label{fig:dblpschema}
 }
   \subfigure[SLAP Dataset]{ 
     \begin{minipage}[l]{0.68\columnwidth}
       \centering
\includegraphics[width=1\textwidth]{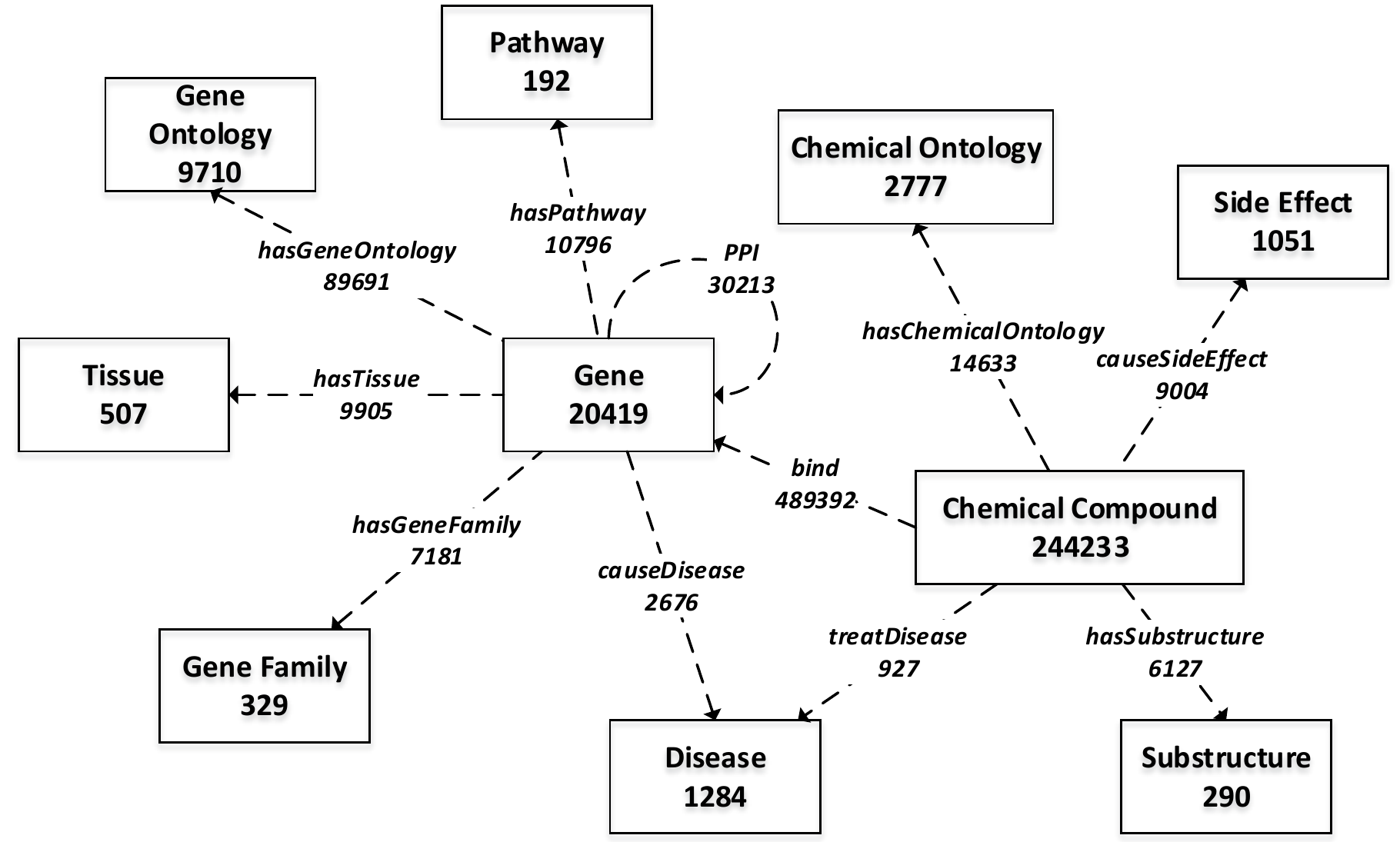}
      \end{minipage}\label{fig:slapschema}
 }
\caption{Schema of datasets} \label{fig:schema}
\end{figure}

\begin{itemize}
\item
\underline{\textbf{ACM Conference Dataset:}} 
	Our first dataset studied in this paper was extracted from ACM
digital library\footnote{http://dl.acm.org/} in June 2011. ACM digital library provides detailed
bibliographic information on ACM conference proceedings, including paper abstracts, citation,
author information {\etc} We extract two ACM sub-networks containing conference proceedings before
the year 2011.
\begin{itemize}
\item 
	The first subset, {\ie}, \emph{ACM Conference-A}, involves 14 conferences in computer
science: SIGKDD, SIGMOD, SIGIR, SIGCOMM, CIKM, SODA, STOC, SOSP, SPAA,  MobiCOMM, VLDB, WWW,  ICML
and COLT. The network schema is summarized in Figure~\ref{fig:acmschema}, which involves five types
of nodes and five types of relations/links. This network includes 196 conference proceedings
({\eg}, KDD'10, KDD'09, {\etc}), 12.5K papers, 17K authors and 1.8K authors' affiliations.  On each
paper node, we extract bag-of-words representation of the paper title and abstract to use as
content attributes. The stop-words and rare words that appear in less than 100 papers are removed
from the vocabulary. Each paper node in the network is assigned with a class label, indicating the
ACM index term of the paper including 11 categories. The task in this dataset is to classify the
paper nodes based on both local attributes and the network information.

\item The second subset, {\ie}, \emph{ACM Conference-B}, involves another 12 conferences in
computer science: ACM Multimedia, OSDI, GECCO, POPL, PODS, PODC, ICCAD, ICSE, ICS, ISCA, ISSAC and
PLDI. The network includes 196 corresponding conference proceedings, 10.8K papers, 16.8K authors
and 1.8K authors' affiliations. After removing stop-words in the paper title and abstracts, we get
0.4K terms that appears in at least 1\% of the papers. The same setups with ACM Conference-A
dataset are also used here to build the second heterogeneous network.
\end{itemize}

\item \underline{\textbf{DBLP Dataset:}}  The third dataset, {\ie}, \emph{DBLP four areas}\footnote{\url{http://www.cs.illinois.edu/homes/mingji1/DBLP_four_area.zip}} \cite{JSDHG10}, is a
bi-type information network extracted from DBLP\footnote{\url{http://www.informatik.uni-trier.de/~ley/db/}}
, which involves 20 computer science conferences and authors. The relationships involve conference-author links and co-author links. On the author nodes, a bag-of-words representation of all the paper titles published by the author is used as attributes of the node. Each author node in the network is assigned with a class label, indicating research area of the author.  The task in this dataset is to classify the author nodes based on both local
attributes and the network information. For detailed description of the DBLP dataset, please refer to \cite{JSDHG10}.

\item \underline{\textbf{SLAP Dataset:}} The last dataset is a bioinformatic dataset SLAP \cite{SLAP}, which is a heterogeneous network composed by over $290K$ nodes and $720K$ edges. 
As shown in Figure~\ref{fig:slapschema}, the SLAP dataset contains integrated data related to chemical compounds, genes, diseases, side effects, pathways {\etc}
The task we studied is gene family prediction, where we treat genes as the instances, and gene family as the labels. In SLAP dataset, each gene can belong to one of the gene family. 
The task of  gene family prediction is that, we are given a set of training gene instances, and for each unlabeled gene instance, we want to predict which gene family the gene belongs to. 
In details,  we extracted $3000$ gene ontology terms (GO terms) and used them as the features of each gene instance.

\end{itemize}

\begin{table}
    \centering
    \caption{Summary of compared methods.}\label{tab:method}
    \begin{tabular}{lllc}
   \toprule
        \textbf{Method}
        & \textbf{Type of Classification}
        & \textbf{Dependencies Exploited}
        & \textbf{Publication}\\
    \midrule
    {\svm} & Multi-Class Classification & All independent &\cite{libsvm}\\ 
    {\ica} & Collective Classification & Citation links & \cite{LG03}\\ 
    {\cp} & Combined Relations & Combine multiple relations & \cite{EN11}\\
    {\cf} & Collective Fusion & Ensemble learning on multiple relations & \cite{EN11}\\
    {\hcc} & Multiple paths & Meta-path based dependencies& This paper\\
    \bottomrule
    \end{tabular}
\end{table}

\subsection{Compared Methods} In order to validate the effectiveness of our collective
classification approach, we test with following methods: 

\begin{itemize}
\item 
 Heterogeneous Collective Classification ({\hcc}): We first test our proposed method, {\hcc},
for collective classification in heterogeneous information networks. The proposed approach can
exploit dependencies based on multiple meta paths for collective classification.\\
\item 
 Homogeneous Collective Classification ({\ica}): This method is our implementation of the ICA
(Iterative Classification Algorithm) \cite{LG03} by only using homogeneous network information for
collective classification. In the homogeneous information networks, only paper-paper links in ACM
datasets and author-author links in DBLP dataset are used.\\
\item 
 Combined Path Relations ({\cp}): We compare with a baseline method for multi-relational
collective classification \cite{EN11}: We first convert the heterogeneous information networks into
multiple relational networks with one type of nodes and multiple types of links. Each link type
corresponds to a meta path in the {\hcc} method. Then, the {\cp} method combines multiple link
types into a homogeneous network by ignoring the link types. We then train one {\ica} model to
perform collective classification on the combined network.\\
\item 
 Collective Ensemble Classification ({\cf}): We compare with another baseline method for
multi-relational collective classification. This method is our implementation of the collective
ensemble classification \cite{EN11}, which trains one collective classification model on each link
types. We use the same setting of the {\cp} method to extract multi-relational networks. Then we
use {\ica} as the base models for collective classification. In the iterative inference process,
each model vote for the class label of each instance, and prediction aggregation was performed in
each iteration. Thus this process is also called collective fusion, where each base model can
affect each other in the collective inference step.\\
\item 
  Ceiling of {\hcc} ({\hcc}-ceiling): One claim of this paper is that {\hcc} can effectively
infer the labels of linked unlabeled instances using iterative inference process. To evaluate this
claim, we include a model which use the \textit{ground-truth} labels of the related instances
during the inference. This method illustrate a ceiling performance of {\hcc} can possibly achieve
by knowing the \textit{true} label of related instances.\\
\item 
  {\hcc} with all meta-paths ({\hcc}-all): Another claim of this paper is that selected meta
path in {\hcc} can effectively capture the dependencies in heterogeneous information networks and
avoiding overfitting. To evaluate this claim, we include a model which uses all possible meta paths
with a maximum path length of 5. This method illustrates the performance of {\hcc} if we
exhaustively involves all possible path-based dependencies without selection.
\end{itemize}

We use LibSVM with linear kernel as the base classifier for all the compared methods.
The maximum number of iteration all methods are set as 10. All experiments are conducted on
machines with Intel Xeon\texttrademark Quad-Core CPUs of 2.26 GHz and 24~GB RAM.

\begin{figure}
  \centering
  \subfigure[ACM Conference-A]{
    \begin{minipage}[l]{0.48\columnwidth}
      \centering
      \includegraphics[width=1\textwidth]{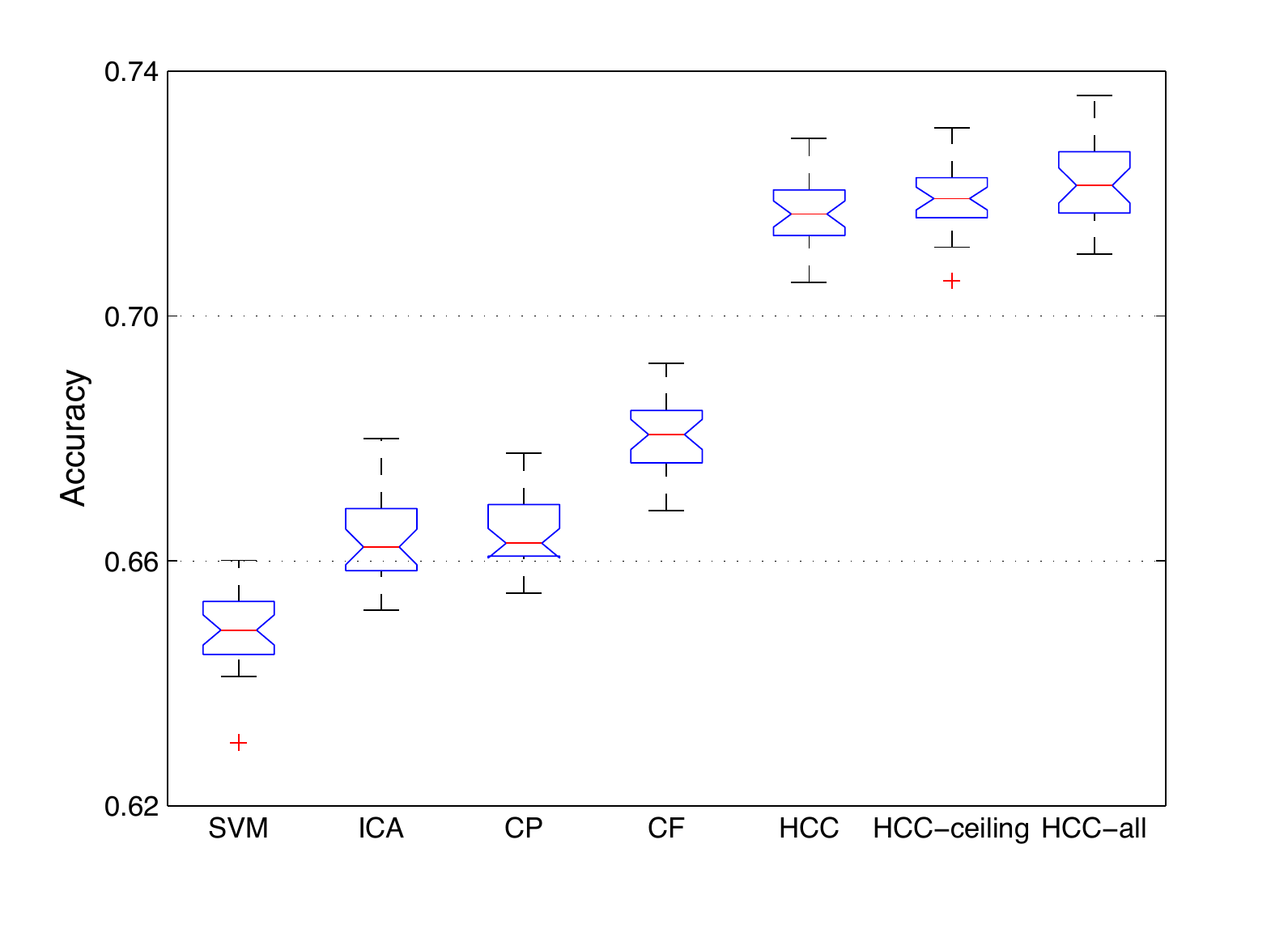}
    \end{minipage}
  }
  \subfigure[ACM Conference-B]{
    \begin{minipage}[l]{0.48\columnwidth}
      \centering
      \includegraphics[width=1\textwidth]{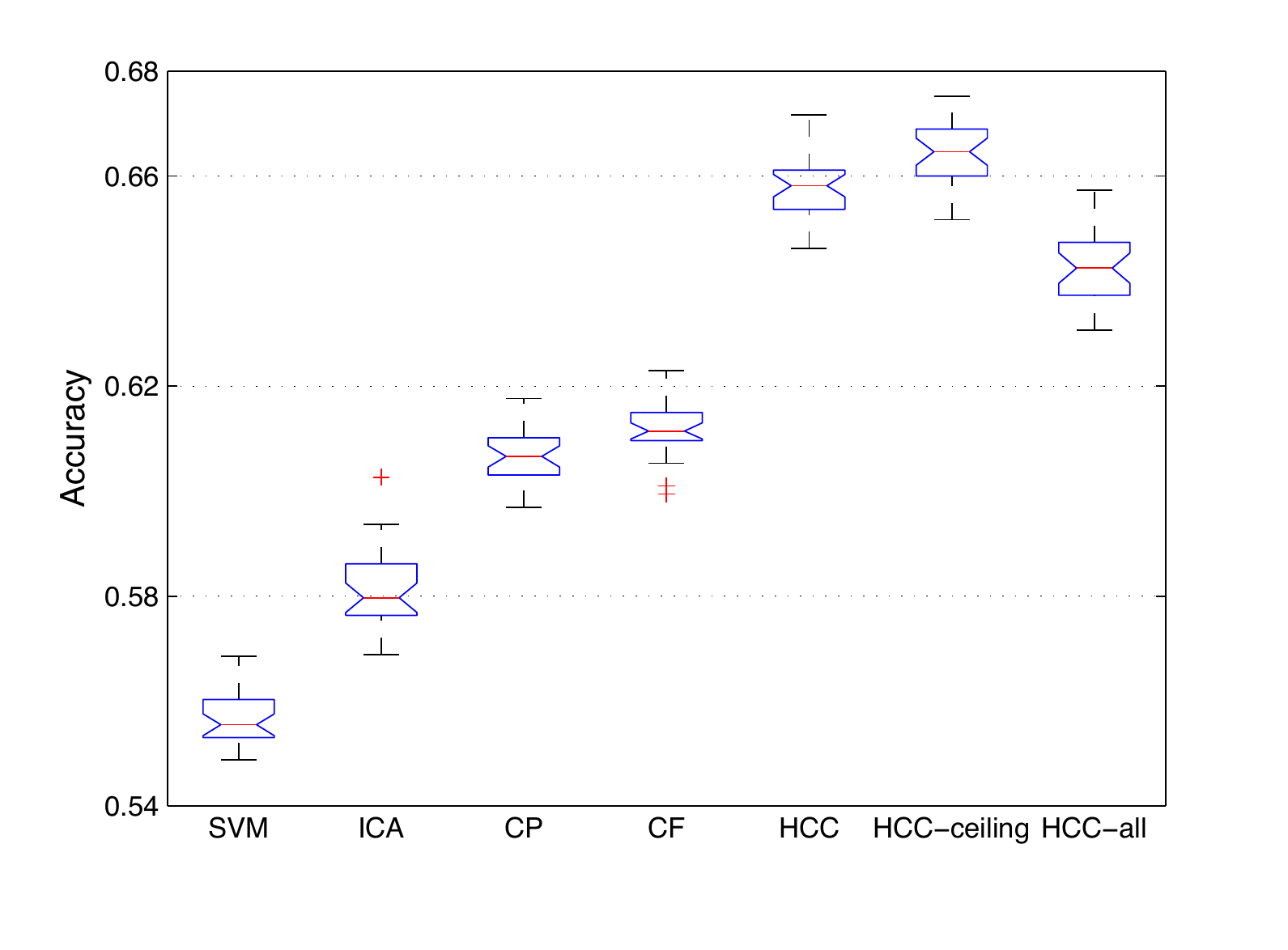}
    \end{minipage}
  }
   \subfigure[DBLP]{
     \begin{minipage}[l]{0.48\columnwidth}
       \centering
       \includegraphics[width=1\textwidth]{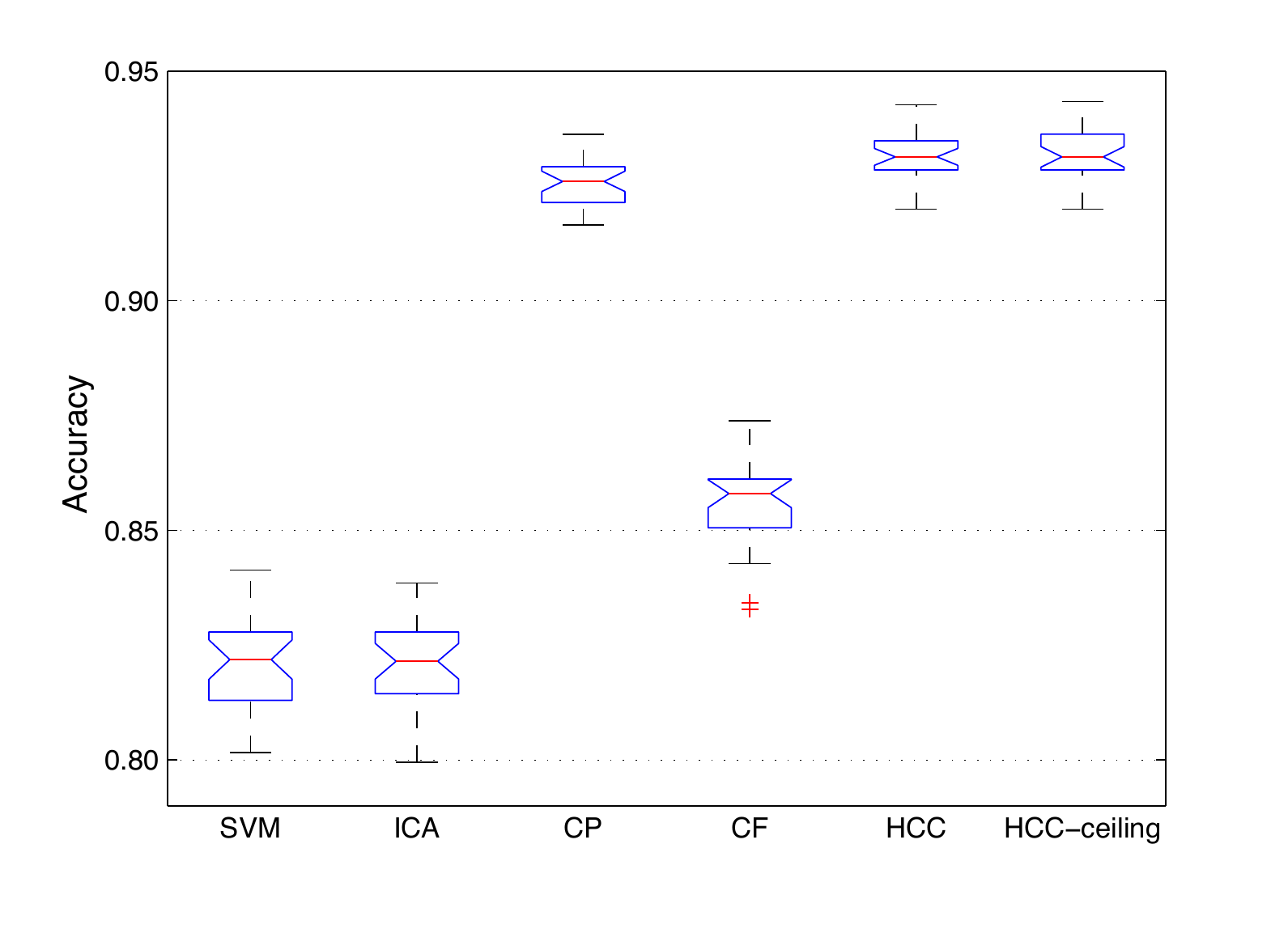}
     \end{minipage}
   }  
   \subfigure[SLAP]{
     \begin{minipage}[l]{0.48\columnwidth}
       \centering
       \includegraphics[width=1\textwidth]{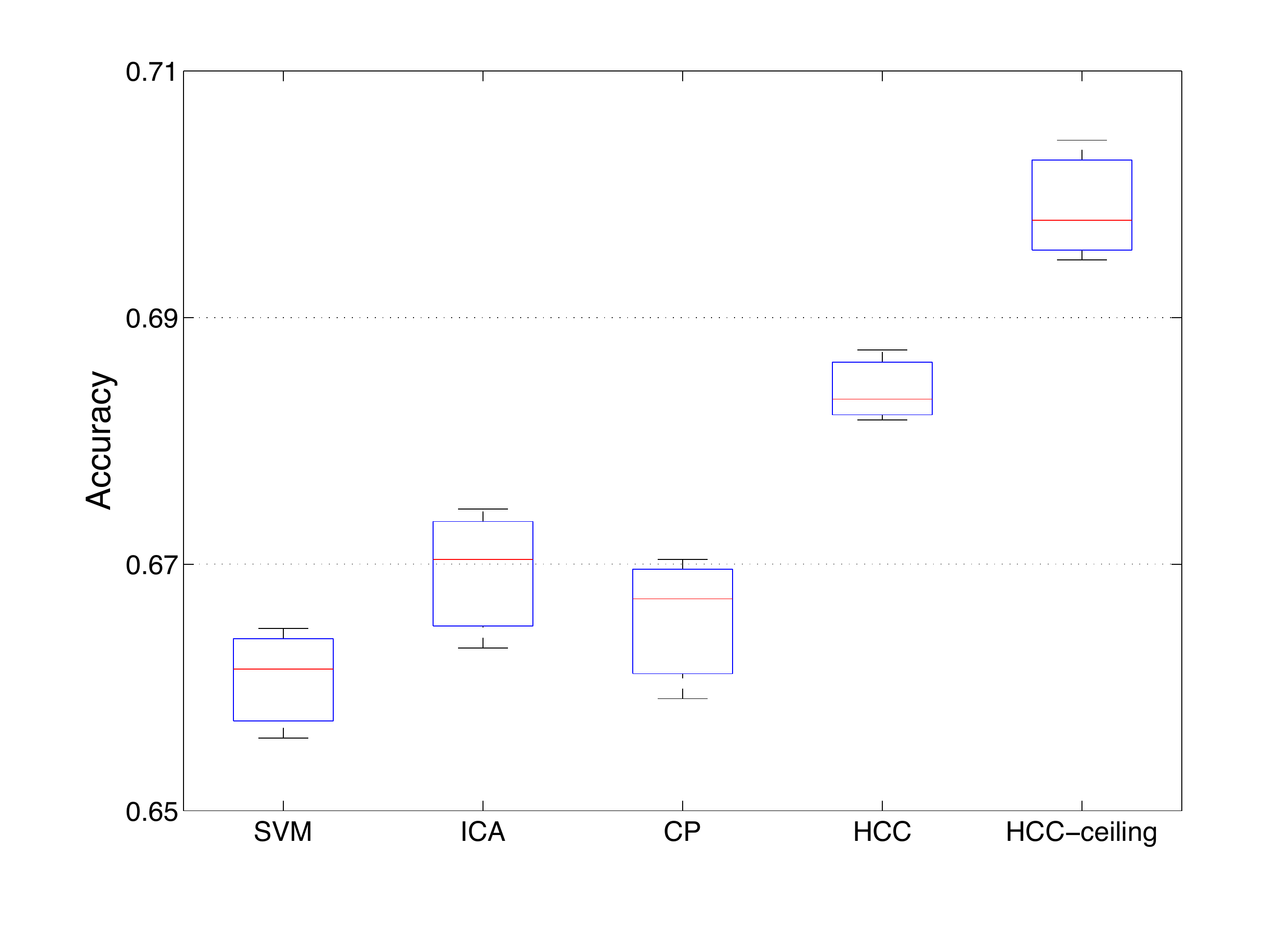}
     \end{minipage}
   } 
   \caption{Collective classification results.} \label{fig:exp1}
\end{figure}

\subsection{Performances of Collective Classification}
\label{sec:PerformanceEvaluation}

In our first experiment, we evaluate the effectiveness of the proposed {\hcc} method on collective
classification.  10 times $3$-fold cross validations are performed on each heterogeneous
information network to evaluate the collective classification performances. We report the detailed
results in Figure~\ref{fig:exp1}. It shows the performances of the six methods on three datasets
with box plots, including the smallest/largest results, lower quartile, median and upper quartile.
In DBLP dataset, note that {\hcc}-all is equivalent to {\hcc} method due to the fact that the schema graph of DBLP is relatively simple.
In SLAP dataset, the schema is much more complex than all the other datasets, and in this case, {\hcc}-all is too computationally expensive. And we didn't show  {\cf} in SLAP dataset, because the performance is not as good as other baselines.

The first observation we have in Figure~\ref{fig:exp1} is as follows: almost all the collective
classification methods that explicitly exploit the label dependencies from various aspects, can
achieve better classification accuracies than the baseline {\svm}, which classify each instance
independently. These results can support the importance of collective classification by exploiting
the different types of dependencies in network data. For example, {\ica} outperformances {\svm} by
exploiting autocorrelation among instances while considering only one type of links, {\ie},
citation links in ACM datasets and co-author links in the DBLP dataset. {\cf} and {\cp} methods can
also improve the classification performances by exploiting multiple types of dependencies. Similar
results have also been reported in collective classification literatures.

Then we find that our meta path-based collective classification method ({\hcc}) consistently and
significantly outperform other baseline methods. {\hcc} can utilize the meta path-based
dependencies to exploit the heterogenous network structure more effectively. These results support
our claim that in heterogeneous information networks, instances can be correlated with each other
through various meta paths. Exploiting the complex dependencies among the related instances ({\ie},
various meta path-based dependencies) can effectively extract the heterogenous network structural
information and thus boost the classification performance. Furthermore, {\hcc} is able to  improve
the classification performance more significantly in datasets with complex network schemas (ACM
datasets) than that with simpler network structure (DBLP dataset).

We further observe that the {\hcc} models perform comparably with the {\hcc}-ceiling models which
had access to the \emph{true} label of related instances. This indicates that the {\hcc} model
reach its full potential in approximated inference process. In addition, {\hcc} method with a small
set of representative paths can achieve also comparable performances with {\hcc}-all models which
includes all meta path combinations with path length $\ell_{max}\le5$. And the performances of
{\hcc} are more stable than those {\hcc}-all in ACM datasets. In ACM Conference-B dataset, {\hcc}
method with fewer meta-paths can even outperform {\hcc}-all method. These results support our
second claim that the heterogeneous dependencies can be  captured effectively by selecting a small
set of representative meta-paths and thus our {\hcc} model can avoid overfitting than using all meta paths.

\begin{table}
\centering
\caption{Results of running time. ``\# path'' represents the number of meta paths explored by the method. }\label{tab:result_time}\vspace{7pt}
\begin{tabular}{llrrrrrrr}
\toprule
dataset &method& \# path  &accuracy& train time (se) &test time (se)& \\
\midrule
\multirow{4}{*}{ACM-Conf-A}& {\svm}	&0	&0.649 &	60.0&25.2& \\
			   &{\ica}			&1	&0.663&	59.9&111.0& \\
      &{\hcc}			&*6	&*0.717& 	*88.3&*440.7& \\
      &{\hcc}-all	&50	&\textbf{0.722}&	332.0&1352.7	& \\
\midrule
\multirow{4}{*}{ACM-Conf-B}& {\svm}			&0	& 0.557& 9.5	&15.2\\
			   &{\ica}			&1 &0.581&16.7&206.6\\
      &{\hcc}			&*6 &*\textbf{0.658}&*36.7&*325.6\\
      &{\hcc}-all	&50 &0.643&202.0&1130.2\\
\bottomrule
\end{tabular}
\end{table}

\subsection{Running Time}\label{sec:result_time}
In Table~\ref{tab:result_time}, we show the running time results of  collective classification methods with different number of meta paths explored. In heterogeneous information networks, there are a large number of possible meta-paths. The more meta-paths are considered in the methods, the training and testing time will both be longer. For example, the {\hcc}-all method incorporates all possible meta paths with path length $\ell \le 5$, {\ie}, 50 meta-paths in ACM datasets.  The training time will be much slower with many meta-paths considered, because the additional computational costs for aggregating class labels through meta paths and the additional dimensions of features for the base learners.  The testing times are also significantly affected by the number meta paths. When more paths are considered, the method needs to aggregate more labels from neighboring instances in each iteration during the inference. Moreover, when more dependencies are considered, the convergence rate of collective classification methods will also be slower. Based upon the above observation, we can see that our {\hcc} method only uses a small set of meta paths, and can achieve comparable or better performances than {\hcc} that uses all meta paths. These results support our motivation on using and selecting meta paths during the collective classification.

\subsection{Influence of Meta Paths}\label{sec:para_path}

In this subsection, we study the influence of different meta paths on the collective classification
performance of our {\hcc} model.  In heterogeneous information networks, different types of meta
path correspond to different types of auto-correlations among instances, thus have different
semantic meanings. In order to illustrate the influence of each path, we compare 6 different
versions of {\hcc} model which exploit different paths separately.

We denote the {\hcc} model with only ``paper-author-paper" path as ``PAP", and it can exploit the
auto-correlation among papers which share authors. Similarly, ``PVP" represents {\hcc} with
``paper-proceeding-paper" path only. Here ``PP*" denotes the {\hcc} with the set of paths that are
composed by citation links: $P$$\rightarrow$$P$, $P$$\leftarrow$$P$,
$P$$\leftarrow$$P$$\leftarrow$$P$, $P$$\leftarrow$$P$$\rightarrow$$P$,
$P$$\rightarrow$$P$$\leftarrow$$P$ and $P$$\rightarrow$$P$$\rightarrow$$P$. In this baseline,
complex paths composed with citation links  ($\ell_{max}=2$) are full exploited.  The ``iid" method
represents the i.i.d. classification model using {\svm}.

\begin{figure}
  \centering
  \subfigure[ACM Conference-A]{
    \begin{minipage}[l]{0.48\columnwidth}
      \centering
      \includegraphics[width=1\textwidth]{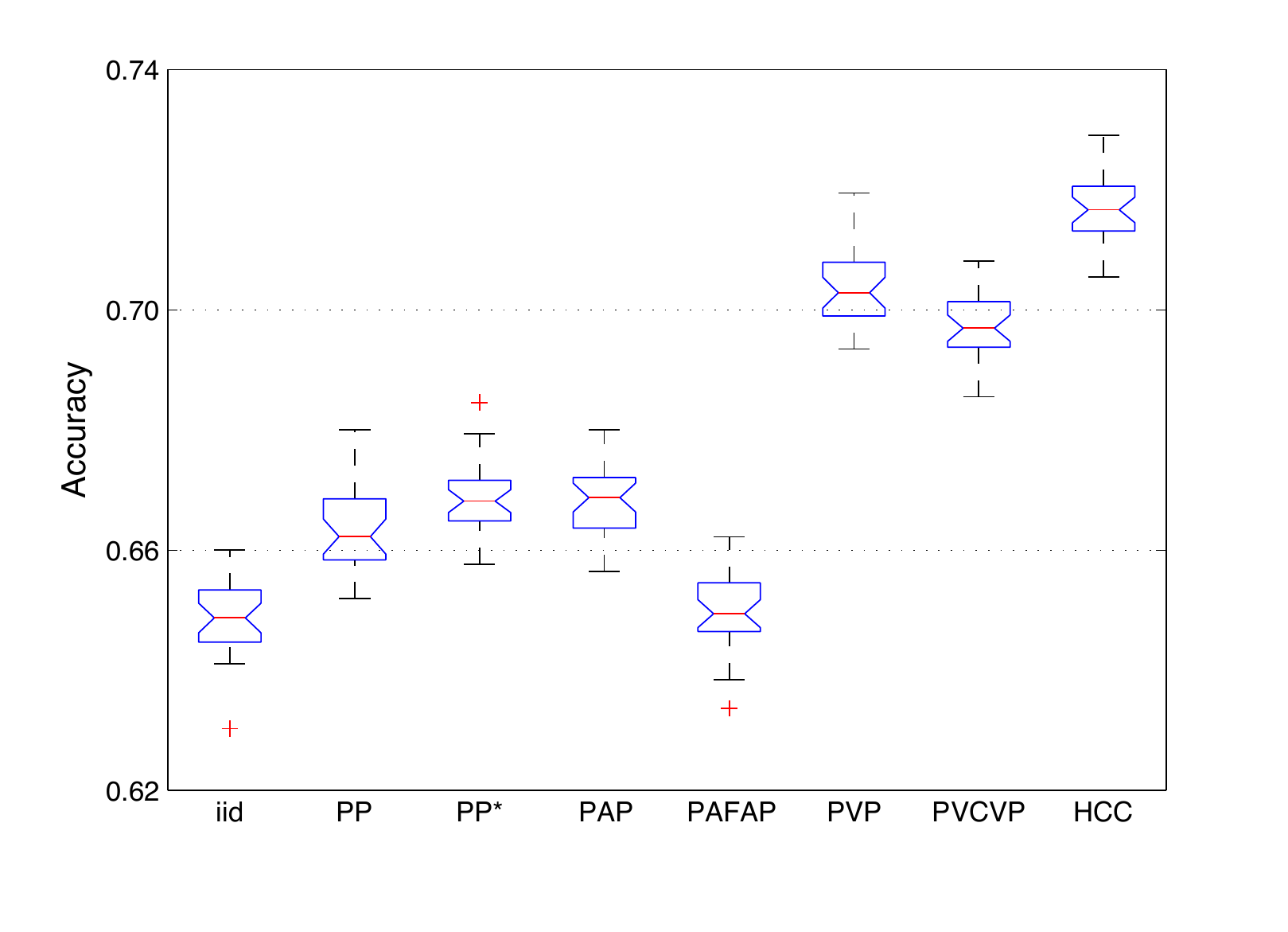}
    \end{minipage}
  }
  \subfigure[ACM Conference-B]{
    \begin{minipage}[l]{0.48\columnwidth}
      \centering
      \includegraphics[width=1\textwidth]{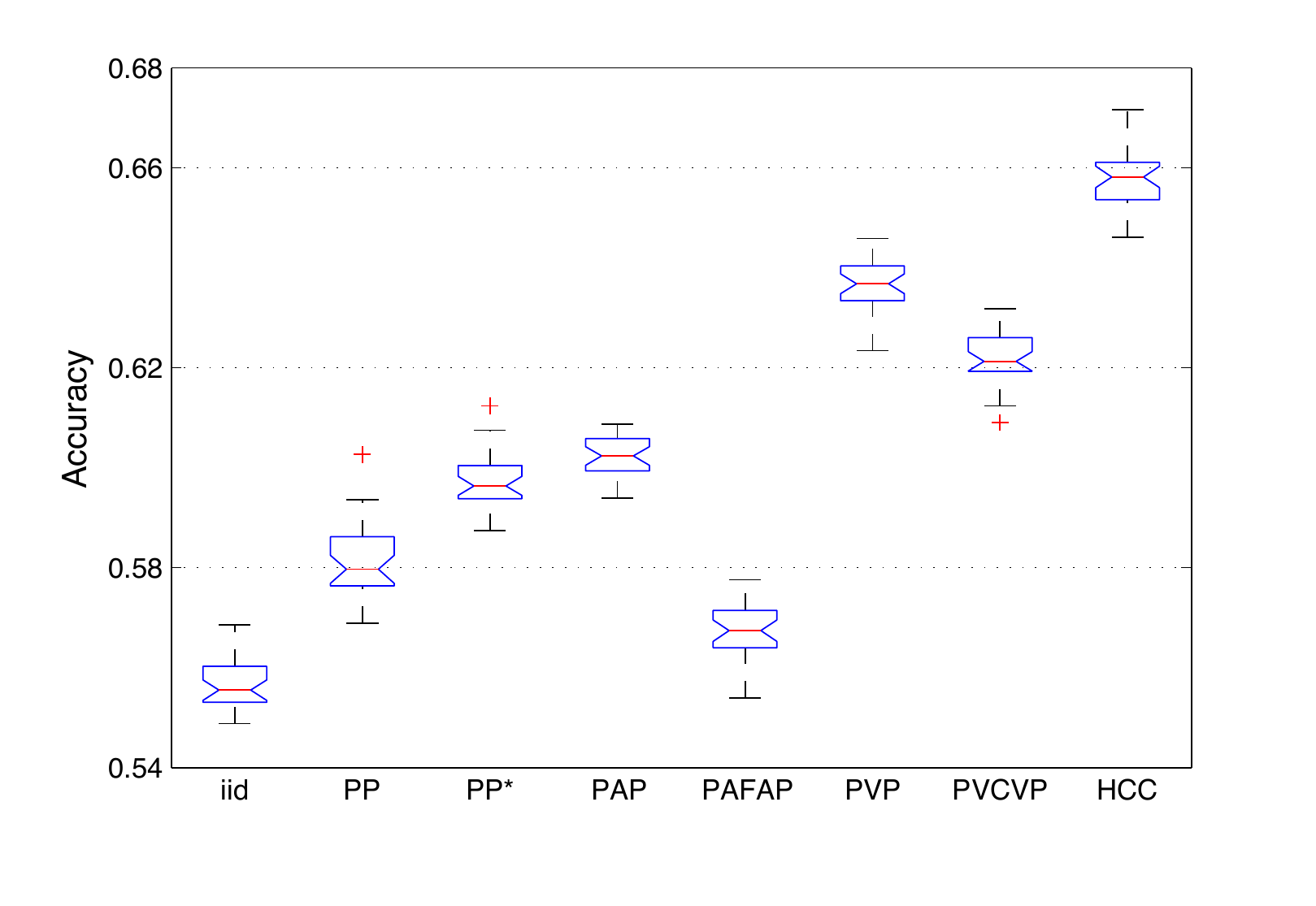}
    \end{minipage}
  }
  \caption{Influences of the meta paths.} \label{fig:exp2}
\end{figure}

Figure~\ref{fig:exp2} shows the collective classification performances using different meta paths.
One can see that two paths are most relevant to the collective classification tasks in ACM dataset:
1) $PVP$: papers in the same proceeding. It indicates that the topics of papers within the same
conference proceeding (also published in the same year) are more likely to be similar from each
other.
2) $PVCVP$: papers published in the same conference (across different years). Since the topics
papers in one conference can slightly changes year after year, but overall the paper topics within
a same conference are relatively consistent. The most irrelevant path is $PAFAP$, {\ie}, papers
from the same institute. It's reasonable that usually each research institute can have researchers
from totally different research areas, such as researchers in the operating system area and those
in bioengineer area.  Moreover, we observe that the performance of PP* that involve different
combination of citation links, such as co-citation relationships can achieve better performances
than $PP$ which only use the citation relationship. This support our intuition that meta path is
very expressive and can represent indirect relationships that are very important for collective
classification tasks.

\section{Conclusion}\label{sec:conclusion}
In this paper, we studied the collective classification problem in heterogeneous information
networks. Different from conventional collective classification approaches in homogeneous networks
which only involve one type of object and links, collective classification in heterogeneous
information networks consider complex structure with multiple types of objects and links. We
propose a novel solution to collective classification in heterogeneous information networks, called
{\hcc} (Heterogeneous Collective Classification), which can effectively assign labels to a group of
interconnected instances involving different meta path-based dependencies. The proposed {\hcc}
model is able to capture the subtlety of different dependencies among instances with respect to
different meta paths. Empirical studies on real-world heterogeneous information networks
demonstrate that the proposed meta path-based collective classification approach can effectively
boost classification performances in heterogeneous information networks.

\bibliographystyle{plain}
\bibliography{hcc_long}

\end{document}